# AGAR: Attention Graph-RNN for Adaptative Motion Prediction of Point Clouds of Deformable Objects

PEDRO GOMES, University College London, United Kingdom
SILVIA ROSSI, Centrum Wiskunde & Informatica, The Netherlands
LAURA TONI, University College London, United Kingdom

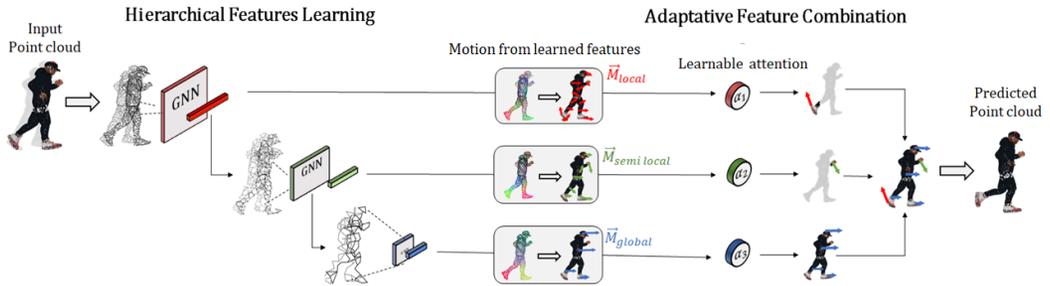

Fig. 1. Proposed deep learning method for motion prediction of dynamic point clouds. Left: By sequentially processing the point cloud at multiple resolutions, the neural network learns hierarchical features. Middle: The hierarchical features correspond to hierarchical motions (from local to global). Right: To predict complex movements, hierarchical motions/features are then combined in an adaptive fashion via learnable weights.

This paper focuses on motion prediction for point cloud sequences in the challenging case of deformable 3D objects, such as human body motion. First, we investigate the challenges caused by deformable shapes and complex motions present in this type of representation, with the ultimate goal of understanding the technical limitations of state-of-the-art models. From this understanding, we propose an improved architecture  for point cloud prediction of deformable 3D objects. Specifically, to handle deformable shapes, we propose a graph-based approach that learns and exploits the spatial structure of point clouds to extract more representative features. Then we propose a module able to combine the learned features in a *adaptive* manner according to the point cloud movements. The proposed adaptive module controls the composition of local and global motions for each point, enabling the network to model complex motions in deformable 3D objects more effectively. We tested the proposed method on the following datasets: MNIST moving digits, the *Mixamo* human bodies motions [14], JPEG [5] and CWIPC-SXR [30] real-world dynamic bodies. Simulation results demonstrate that our method outperforms the current baseline methods given its improved ability to model complex movements as well as preserve point cloud shape. Furthermore, we demonstrate the generalizability of the proposed framework for dynamic feature learning, by testing the framework for action recognition on the MSRAction3D dataset [17] and achieving results on-par with state-of-the-art methods.

## 1 INTRODUCTION

Point cloud sequences are a flexible and rich geometric representation of volumetric content used in a wide range of applications from autonomous driving [23, 26], robotics [20, 49] to virtual/mixed-reality services [4, 32]. Such sequences consist of consecutive point clouds, each composed of an unordered collection of 3D points representing 3D scenes or 3D objects. Although the point cloud is a highly appealing representation impacting multiple sectors, how to properly process it is still an open challenge. One of the most successful methodologies was the development of neural networks able to learn directly from unstructured point cloud data. This approach was pioneered by PointNET [28] architecture, which learns features by processing each point independently. However, in such architecture, the local structures within the point cloud are neglected. This is a strong limitation since local structures contain key semantic information about the 3D geometry. To address this issue, PointNet++ [29] introduced a point-based hierarchical architecture that considers hierarchical neighbourhoods of points rather than acting on each of them independently. The network processes point neighbourhoods at increasingly larger scales along a multi-resolution hierarchy, as shown on the left side of Fig. 1. This approach groups the local features learned from small neighbourhoods into larger units and processes them to learn higher-level features, allowing the network to abstract the multiple structures within the data. Although PointNet++ hierarchical architecture was initially designed for static point clouds, it has since been extended to the case of dynamic point clouds [18, 19]. In these cases, instead of extracting features from neighbourhoods in a single point cloud, the network extracts dynamic features from a hierarchy of spatio-temporal neighbourhoods across time. The learned dynamic features can be applied to a wide range of downstream tasks, such as action classification, motion prediction and segmentation. In this paper, we focus on the point cloud prediction task. Specifically given a point cloud sequence $\mathcal{P} = \{P_1, \ldots, P_T\}$, composed of $T$ frames with $p_{i,t} \in \mathbf{R}^3$ being the Euclidean coordinates of point $i$ in point cloud $P_t \in \mathbf{R}^{N \times 3}$, our goal is to predict the coordinates of future point clouds $(\hat{P}_{T+1}, \ldots, \hat{P}_{T+Q})$, where $Q$ is the prediction horizon.

At the moment, point-based hierarchical methods can be considered the de-facto state-of-art approach for point cloud prediction. However, while these methodologies have shown good performance when predicting simple and rigid movements as translations in automobile scenes [3], they are often limited when predicting the motion of 3D deformable objects. Addressing this limitation is the main goal of this paper. Predicting deformable objects is challenging since the point cloud shape changes over time and performs highly *complex* motions. For example, in a 3D representation of a football player running or a dancer performing during a music event, their point cloud representations change over time following different postures. Moreover, the performed movements are not rigid transformations but rather a combination of multiple and diverse local motions. For instance, if we imagine the player rising the hand while running, their arm and hand will be characterised by a combination of movements (i.e., local rising movement and the global forward translation). Given their characteristics, processing 3D deformable objects presents two major challenges: (i) establishing point correspondence across time and preserving the shape of the predicted point cloud; (ii) generating accurate motion predictions that are a composition of multiple movements at different levels of resolution.

To address the above challenges, we must first understand if the current state-of-art models are able to address such challenges. Within this context, we first demonstrate the model's inability to establish precise temporal correlations and preserve the predicted point cloud shape. This is because they fail to consider the structural relationships between the points during the learning process. Then, to investigate the challenge of predicting complex motions, we employ the explainability techniques introduced in our previous work [12]. These techniques demonstrated that the hierarchy of dynamic features corresponds to learning from local to global motions (in the centre of Fig. 1). In this paper, we build upon this interpretation to identify the technical limitations of the current framework approach. Specifically, we show that in most methodologies [6, 11, 18, 25] to generate predictions of future motions the hierarchical features are combined via learnable weights. Most critically to preserve permutation invariance, when combining hierarchical features, the same learned weights are applied to all points across frames. However, in deformable objects, not all points benefit from the same combination of hierarchical features. For example, some points can be described entirely by global motions, while other points are better described by a combination of global and local motions. We show that this *fixed* combination of hierarchical features is a key limitation to the network's ability to predict complex motions.

Based on the limitations identified above, we propose AGAR: an attention-based hierarchical graph-recurrent neural network (RNN) for point cloud prediction of deformable objects. Our proposed architecture includes an initial graph-based module that extracts the underlying geometric structure of the input point cloud as spatial features. From the learned spatial features, we construct a *spatio-temporal graph* that forms more representative neighbourhoods than current methods that neglect the point cloud structure. The graph is then processed by sequential graph-RNN cells that take structural relations between points into account to learn dynamic features. To address the limitation of the fixed combination of hierarchical features, we propose a novel module denoted as *Adaptative feature combination*. The proposed module employs an attention mechanism to dynamically assign different degrees of importance to each level of hierarchical features. As such, for each point, the network can control the composition of the local and global motions that best describe the point behaviour. This concept is illustrated in the right part of Fig. 1, where the network selects the regions that benefit from particular motions (i.e. local, semi-local, global) instead of blindly combining all the motions learned in the multiple hierarchical levels. Besides improving the prediction of complex motions, the *Adaptative feature combination* module, is also an explainability tool. The module allows us to visualize the influence of each learned feature on the predicted motion, providing a deeper understanding of the network's internal workings.

The proposed method is trained in a self-supervised fashion and it is tested on several datasets such as the *Mixamo* synthetic human bodies activities dataset [14], JPEG [5] and CWIPC-SXR [30] real-world human bodies datasets and compared against state-of-art methods. To extend such comparison, we also tested on a dataset of rigid objects (moving MNIST point cloud dataset [6, 31]), and a dataset of automobile scenes (Argoverse dataset [3]). A powerful strength of our framework is the ability to extract the general dynamic behaviour of the point cloud as dynamic features. Since such features are useful for downstream tasks, we also tested the proposed architecture for the action recognition task on human bodies (MSRAction3D dataset [17]). The proposed method outperforms the state-of-art methods in human bodies prediction and achieves on-par results for rigid objects and automobile scene prediction as well as for the action recognition task. The results demonstrated that our proposed method can leverage the structural relations between points to learn more accurate representations and preserve the point cloud shape during prediction. The results further show that the proposed *Adaptive feature combination* module predicts complex motions in human bodies with more accuracy than the current state-of-art approaches. Lastly, the code and datasets required to reproduce the work are made publicly available at https://github.com/pedro-dm-gomes/AGAR.

In summary, the key contributions of our work are:

- Understanding of current state-of-the-art frameworks key limitation for generating motion flow prediction. We show how the current approach is equivalent to combining learned local and global motions without regard to the point position in space and time, and how this strategy fails to model the complex motions present in deformable objects.
- A novel module that combines hierarchical features in an adaptive manner according to the scene context. The proposed module dynamically controls the composition of local and global motions for each point, allowing the network to predict complex motions with higher accuracy and flexibility. This also offers an explainability tool.
- A graph-based module that exploits the point cloud geometric structure to form spatio-temporal neighbourhoods from where the meaningful dynamic features can be extracted. The structural information is further included in the learned dynamic features, reducing the deformation of the predicted point cloud shape.

The remainder of this article is organized as follows. In Section 2, we provide a state-of-art of research for point cloud prediction. In Section 3, we study the hierarchical component and identify the limitations of the state-of-art prediction framework. Based on the limitations identified, in Section 4, we propose AGAR, an improved architecture with graph-RNN cells and a novel *Adaptive feature combination* module. Section 5 describes implementation details. Finally, the experimental results and conclusion are presented in Section 6 and Section 7, respectively.

## 2 BACKGROUND

This section provides an overview of the research in dynamic point cloud processing (Section 2.1), followed by a detailed description of the current state-of-art point cloud prediction framework and the notation used throughout this paper (Section 2.2).

### 2.1 Related Works:

In the current literature, dynamic cloud processing has been approached from multiple overlapping directions related to motion prediction (e.g., segmentation, and action recognition). These high-level tasks share a common challenge: extraction of temporal correlations between sequential point cloud frames, challenged by the irregular structure and by the lack of explicit point-to-point correspondence across time. In the following, we summarize the approaches proposed in the literature to overcome such challenges and how they lead to the development of the current state-of-the-art framework for point cloud prediction.

An initial approach to learn from irregularly structured data such as point cloud data was to convert them into a regular representation such as 2D multi-view [9, 27, 50] or 3D voxels [22, 38, 42] and then process the converted data with traditional neural networks. This approach, however, suffered from high memory consumption and quantization errors. Within this context, the hierarchical architecture proposed in PointNet++ [29], able to process raw point cloud data directly, has become a pillar of work for learning-based point cloud processing. The PointNet++ hierarchical architecture has been extended to dynamic point clouds, by introducing spatio-temporal neighbourhoods to extract dynamic features. The spatio-temporal neighbourhoods still lack explicit point-to-point correspondence over time. However, by processing the neighbourhoods at multiple scales, the network can capture temporal correlations that would otherwise be hidden. This hierarchical learning strategy has proved to be highly successful at learning from point cloud sequences and has been widely adopted throughout the literature [8, 10, 15, 18, 19, 24, 37, 45]. In PSTNet [8] a hierarchical architecture is used for the action classification of point cloud sequences. In PointPWC-Net [45] a hierarchical architecture learns motion in a course-to-fine fashion by learning a motion flow and a cost function between two adjacent frames at each hierarchical level. More recently, attention-based mechanisms have been incorporated into hierarchical architectures [7, 35, 36, 39, 41]. The use of attention allows the network to selectively focus on the most important parts of the point cloud. Although attention mechanisms do not fully address point-to-point correspondence, they allow for a more flexible construction of hierarchical neighbourhoods by enabling selective aggregation within the network. For example in [36], an attention mechanism is used to sample the most critical points, enabling the network to better correspond points over time. In [7], attention is incorporated into the spatio-temporal point aggregation, assigning greater weight to points that are more similar to the target point during the feature aggregation. It is worth noting, these aforementioned attention-based works, learn the attention of a point relative to the features of other points, with the goal of improving the extracted features. We, on the other hand, propose to learn the attention of a point relative to the features of each hierarchical level with the goal of refining the predicted motion.

Although the methods presented above have demonstrated their ability to extract features from point cloud sequences, they suffer from several drawbacks when specifically applied to the point cloud prediction task, which is the focus of this paper. For instance, methods such as PointPWC-Net [45] learn a motion flow between two adjacent frames instead of learning a future motion to predict the next frames, preventing the model from capturing long-term movements. Other methods such as PSTNet [8] are able to capture long-term correlations by processing all the sequence frames simultaneously. While this is an effective approach for classification or segmentation tasks, the memory required to process all the frames simultaneously prevents this approach to be scaled to long sequences or applied to iterative prediction tasks. These drawbacks led to the implementation of point-based hierarchical architectures into RNNs or their variants, e.g., Long Short-term Memory (LSTM) and Gated Recurrent Unit (GRU). These types of models are designed to model sequential data taking only one frame as input at each interaction. The key characteristic of RNNs is their hidden states that can act as a *memory*. The states store information from prior inputs and are continuously updated. As a result, the output of RNNs depends on the input but also the prior elements within the sequence. Pioneer in this framework, PointRNN [6] learns dynamic features from spatio-temporal neighbourhoods between two adjacent frames. The learned dynamic features are then used as states storing the point's history of movements and used to learn the next interaction features as well to predict future movements. This methodology inherits the ability to capture the long-term dynamic behaviour of sequential data from RNN models while having low-memory requirements. Following this approach, several works [6, 21, 43, 44, 47, 48] combined RNN cells or its variants into hierarchical architectures to model point cloud sequences. These point-based hierarchical RNN architectures are currently the state-of-art approach for iterative point cloud prediction. However, the majority of current methods proposed in the literature are focused on predicting the motion of point clouds from rigid objects, leaving the unique challenges associated with predicting point clouds from deformable objects overlooked. Our work aims to address this gap, by first identifying the current challenges caused by such objects and by developing models specifically designed to handle such challenges.

## 2.2 Hierarchical Point-based RNN Architecture for Point Cloud Prediction

In this section, we present an architecture that characterizes the state-of-art hierarchical RNN framework used for point cloud prediction. We will use this model to identify the key challenges of current state-of-art (Section 3) and to highlight the novelty of the solutions proposed in this paper (Section 4). Table 1 summarizes the main notation used throughout the paper. Without loss of generality, we describe the iterative prediction framework depicted in Fig. 2. Given a point cloud sequence $\mathcal{P}$, at each interaction, the network processes one input point cloud $P_t \in \mathbf{R}^{N \times 3}$ and outputs the prediction of the point cloud at the next time step $\hat{P}_{t+1}$. The framework can be described by three main phases:

(1) Dynamic Extraction (DE) phase: the network processes the input point cloud $P_t$ and extracts the point cloud dynamic as multiple $L$ levels of hierarchical features $(D_t^1, ..., D_t^L)$.
(2) Feature Propagation (FP) phase: combines the learned features from multiple levels into a single final dynamic feature $D_t^{\text{Final}}$;
(3) Prediction phase: The final features are converted via a fully-connected layer into motion vectors $M_t$ and added to the input point $P_t$ cloud to predict the point cloud $\hat{P}_{t+1}$ at the next time step.

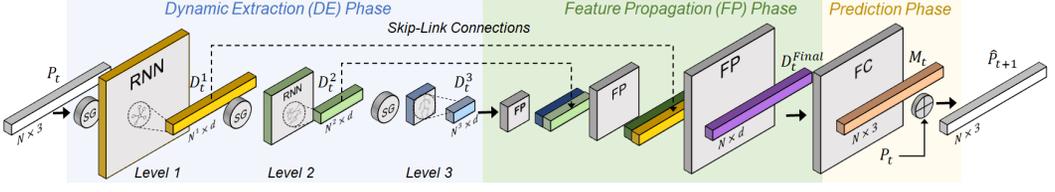

Fig. 2. **Generic state-of-art framework for point cloud prediction** for interaction at time $t$. The architecture is composed of a Dynamic Extraction (DE) phase, Feature Propagation (FP) phase and a prediction phase.

We now describe the DE and FP phases in more detail. Being straightforward we omit the detailed description of the prediction phase.

*2.2.1 Dynamic extraction (DE) phase.* Depicted on the left part of Fig. 2, the DE phase consists of multiple sequential RNN cells, for a total of $L$ levels (in the figure $L = 3$). Before being processed by each RNN cell, the point cloud is downsampled by a *Sampling and Grouping* (SG) module, as described in [29]. At each RNN cell, for each point, a dynamic feature is extracted by aggregating information from the point spatio-temporal neighbourhood. In the majority of methods [18, 21, 25] the neighbourhood of each point is defined as the $k$ nearest neighbour ($k$-nn) points in the previous frame, where the proximity is measured using the Euclidean distance between point 3D coordinates. The RNN cells are sequentially stacked in order to have the dynamic features learned at an RNN cell be the input of the next RNN cell. It is worth noting that the subsequent sampling, which results in a sparser point cloud at later levels/RNN cells, is responsible for the creation of hierarchical neighbourhoods with a progressively larger geometric distance between points. Thus, the first level ($l = 1$) learns local dynamic features $D_t^1$ from small-scale neighbourhoods, whereas the last level $l = L$ learns global dynamic features $D_t^L$ observing large-scale neighbourhoods.

| Terminology | Description |
| --- | --- |
| level | network layer extracting dynamic features at a specific resolution. |
| spatial features | vectors describing the point's local geometric structure. |
| dynamic features | vectors describing the point's dynamic behaviour. |

| Parameter | Description |
| --- | --- |
| $\mathcal{P}$ | sequence of point clouds. |
| $T$ | number of points clouds (frames) in sequence. |
| $l, L$ | level and the total number of levels. |
| $N, N^l$ | original number of points and number of points at a level $l$. |
| $P_t^l \; p_{i,t} \in P_t$ | point cloud and cartesian coordinates of point $i$ |
| $\hat{P}_t^l \; \hat{p}_{i,t} \in \hat{P}_t$ | predicted point cloud and cartesian coordinates of predicted point $i$ |
| $k$ | number of point neighbourhoods. |
| $S_t^l \; s_{i,t}^l \in S_t^l$ | point cloud spatial features and spatial feature of point $i$. |
| $D_t^l \; d_{i,t}^l \in D_t^l$ | point cloud dynamic features and dynamic feature of point $i$. |
| $M_t, m_{i,t} \in M_t$ | point cloud motion vectors and motion vector of point $i$. |
| $D_t^{\text{fp}} \; d_{i,t}^{\text{fp}} \in D^{\text{fp}}$ | dynamic features propagated from level $l$ to $l-1$. |
| $D_t^{\text{Final}} \; d_{i,t}^{\text{Final}} \in D^{\text{Final}}$ | point cloud final dynamic features and final feature of point $i$. |
| $\Theta_{\text{FP}}, \Theta_{\text{S}}, \Theta_{\text{D}}, \Theta_{\text{R}}, \Theta_\alpha$ | learnable network weights. |
| $G_t^{\text{C}}$ | coordinate graph. |
| $G_t^{\text{ST},l}$ | spatio-temporal graph. |
| $m_{ij}^l$ | message vector for node $j$ to node $i$. |
| $\alpha_i^l$ | attention value of point $i$ to the feature of level $l$. |

Table 1. Terminology & Notation

*2.2.2 Feature Propagation (FP) phase.* Once the DE phase has learned the features from all the levels ($D_t^1, ..., D_t^L$), the FP phase combines them into a single final feature ($D_t^{Final}$). Currently, the most popular architecture for features combination is the original architecture proposed in PointNet++[29], which is also found in most state-of-art methods without significant differences. We will refer to this architecture as state-of-art *Classic-FP* (depicted in the green side of Fig. 2). In the *Classic-FP* the features combination is done by hierarchically propagating the features from the higher levels to the lower levels using several FP modules [29]. At each module, the sub-sampled features from the higher level are first interpolated to the same number of points as the lower level. The interpolation is done by weighted aggregation of the features of the three closest $j$ points in the sub-sampled point cloud as such:

$$\tilde{d}_{i,t}^l = \frac{\sum_{j=1}^3 dist_{ij,t} \times d_{i,t}^{l+1}}{\sum_{j=1}^3 dist_{ij,t}}, \quad dist_{ij,t} = \frac{1}{||p_{i,t}^l - p_{j,t}^{l+1}||^2} \tag{1}$$

where $\tilde{d}_{i,t}^l \in \tilde{D}_t$ is interpolated features from the number of points at level $l + 1$ to the number of points at level $l$. The interpolated high-level features are then concatenated with a skip-linked connection to lower-level features at the same number of points. The concatenation is processed by a point-based network that processes each point independently via shared weights $\Theta_{FP}^l$ as follows:

$$D_t^{l^{FP}} = ReLU\left(\Theta_{FP}^l \{D_t^l; \tilde{D}_t^l\}\right). \tag{2}$$

The process is repeated in a hierarchical manner until the features from all the levels have been combined into final features ($D_t^{Final}$).

# 3 CHALLENGES AND LIMITATIONS

The hierarchical point-based RNN framework, presented in the previous section, suffers several limitations when facing the challenge of processing deformable objects such as human-body-like sequences. In this paper, we explain why those challenges arise and how to overcome them. In the following, we disentangle the challenges of current models as *i)* challenges in processing/predicting objects with deformable shapes (Section 3.1); *ii)* challenges in predicting complex motions (Section 3.2). Taking advantage of the understanding built in this section, in Section 4 we introduce our proposed method, built to overcome the main limitations identified here.

## 3.1 Challenges in Processing Deformable Shapes

The main challenges encountered in processing and predicting objects with deformable shapes, such as clothing, food, or human bodies are *i)* having a semantically-meaningful point-to-point correspondence (used to learn dynamic features); *ii)* avoiding shape distortion (which is highly noticeable in 3D objects and therefore of high-negative impact on cloud prediction quality).

The challenge of establishing point-to-point correspondence is present in any point cloud processing, but it is clearly exacerbated in the case of deformable 3D objects. The majority of current works follow the same strategy as PointRNN [6] and assume that the points in the current frame are matched with points in close proximity in the previous frame. This proximity is built in the 3D Euclidean space. However, in 3D deformable objects, points that are geometrically close in space are not necessarily semantically correlated and not necessarily belong to the same segment of the object. Fig. 3, shows three examples of how matching based on geometric proximity can lead to the creation of misleading neighbourhoods. This means that point correspondence across time is challenged by the mismatch between Euclidean proximity and semantically-meaningful proximity.

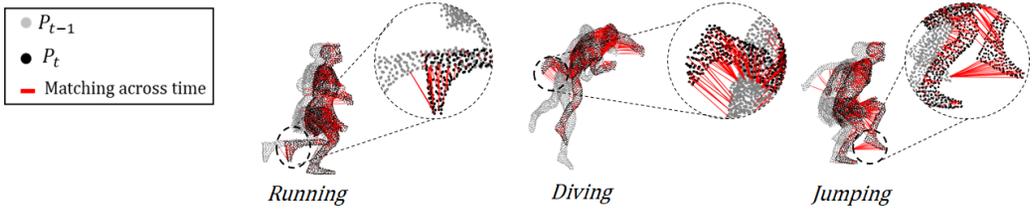

Fig. 3. Example of matching points across time using geometric coordinates for three sequences: *Running*, *Diving*, *Jumping* from [14] (These sequences are examples of particularly high motions for visualization purpose). The dashed circles show a zoom-in of the regions where grouping using coordinates would create incorrect neighbourhoods. For example in the *Running* sequence, the points in the foot at time $t$ are incorrectly matched with the points in lower-leg at $t-1$.

On the other hand, current methods often struggle to preserve the predicted point cloud shape. This is mainly due to the fact that a separate motion vector is learned for every point with no clear semantic constraints. If these motion vectors vary significantly among neighbouring points, the result is a prediction with a deformed shape. This issue can be tackled by imposing *hard* shape constraints, such as learning a single motion vector for all the points in a region. However, this strategy can only be applied to rigid objects. In deformable objects, the object shape changes according to different postures, meaning points must be allowed to have separate motions. Thus, it is important to strike a balance between preserving the shape and having enough per-point motion flexibility to predict possible shape variations. The key to achieving this balance is to capture the underlying semantic structure and take it into account as a soft shape constraint during the learning process.

Both challenges of point correspondence and shape deformation can be summarized in the following limitation: **Lack of structural relationship between points in point cloud prediction (Limitation 1)**. Learning and exploiting this prior in the learning process is one of the novelties of our proposed model and it will be specifically addressed by learning a semantically-meaningful graph and exploiting this graph when extracting features (via graph-RNN cell).

## 3.2 Challenges in Processing Complex Motions

A second key challenge present in processing 3D dynamic objects as the human body is that the movement of such objects is usually a *complex motion*. Complex motions refer to movements that involve a combination of multiple degrees of freedom such as translation, rotation, and deformation, which are applied to different parts of the object independently. This is typical of deformable objects or any 3D objects with disjoint components, each of them with its own movement. As an example, consider a point cloud representing a human body running forward (Fig.4 (a)-*Man-Running*). While the full body moves forward (translation), the person swings their arms (rotation), and their hand bends from an open to a closed position (shape change). The complex nature of such movements makes them challenging to be accurately captured and predicted. Based on a novel visualization technique that we introduced in our previous work [12] on explainability, we now highlight key limitations of the current architectures. Specifically, we show how complex motions can be seen as a sum of low-, medium- and high-level motion leading to an understanding that the current model suffers from the following main limitation: **the fixed combination of hierarchical features in the prediction phase (Limitation 2)**. We now explain this limitation in more detail.

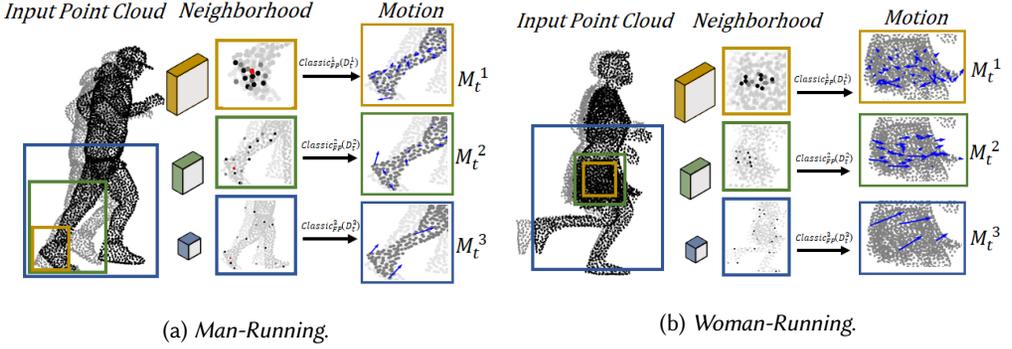

(a) *Man-Running.*          (b) *Woman-Running.*

Fig. 4. Hierarchical of dynamic features as motion vectors given for two input sequences (*Man-Running* and *Woman-Running*). For each sequence, the figure shows input dynamic point cloud; multi-scale neighbourhood at different levels; motion vectors learned at each level of the network.

In our explainability work [12], we have demonstrated that motion vectors inferred by hierarchical architectures (Fig.2) can be disentangled into individual motion vectors produced at each hierarchical level, as follows.

$$M_t^l = Classic_{FP}^l(D_t^l) \tag{3}$$

$$M_t = \sum_l^L M_t^l \tag{4}$$

where, $Classic_{FP}^l$ is the function that replicates the operation of the *Classic-FP* in a disentangle manner converting the learned feature at each level $l$ to an individual motion vector $M_t^l$, and $M_t$ is the final predicted motion vectors outputted by the network. This leads to the interpretation that current approaches in the literature **model complex motions as a combination of local and global motions**, which are learned as hierarchical dynamic features. This is illustrated in Fig. 4 which depicts the dynamic features as motion vectors and the hierarchical neighbourhoods given two point cloud sequences as input to a state-of-art prediction architecture (presented in Fig.2) with three levels ($L = 3$) [12]. In both sequences, it can be seen that the lower level learns features only by looking at points in a small area (top gold squares in the figure). In contrast, the higher level learns features by considering a sparser set of points in a large area (bottom blue squares in the figure). In the example in Fig.4 (a), in which the runner's foot performs a complex motion, it can be observed that the lowest level captures small and diverse motions (e.g., rotation of the heel) $M_t^1$, while the highest level learns the forward motion of the entire body $M_t^3$.

This interpretation of features as motion vectors can be generalized for the majority of current methods because while they differ in the feature extraction process, they all share the *Classic-FP* strategy to perform the motion reconstruction process. As such we elaborate on this explainability technique to identify current state-of-art framework limitations to predict complex motions. Namely, the motion vector prediction is obtained by combining the dynamic features from the different levels via a learned weighted combination. However, each point motion is obtained using the same set of combination weights $[\Theta_{\text{FP}}^1, \ldots, \Theta_{\text{FP}}^L]$ for *all* points, frames, and sequences. As a result for every point, regardless of its position space and time, the predicted motion is obtained by the same fixed combination of local, medium and global motions. Based on this technique we can understand that *i)* different features can be associated with the different levels of motions forming the complex resultant motion, and *ii)* knowing different parts of the objects might be subject to different types of movements highlights the strong limitation in having the same combination of motion levels. Specifically, while a set of weights might lead to the appropriate combination of the motion vectors in Fig.4 (a), in which a local movement is analysed (foot), it does not hold in the case of the *"Woman-Running* sequence in Fig.4 (b), in which a more global movement is highlighted (torso). The points in the lower torso perform a rigid movement forward corresponding to the global motion of the body, while the lower part of the body performs a quite dynamic rotation of the foot. This means that only the global motion vector (pointing forward) would be sufficient to describe the movement of the torso. However local features (hence local motions) cannot be neglected, since this would lead to neglecting the local motions in parts with strong local movement such as the foot. As a result, in Fig. 4 (b) local motion vectors ($M_t^1$) clearly lose any motion interpretation and becomes instead random vectors mainly used to compensate for the erroneous addition of multiple motion vectors in this part of the body.

It is worth mentioning that while this understanding might appear straightforward, to the best of our knowledge, this is the first work explaining PointRNN and similar hierarchical architectures when processing 3D deformable objects, showing the limitation in adopting a fixed combination of hierarchical features in the prediction phase. In the next section, we propose an architecture that overcomes this limitation by introducing an attention-based mechanism in the prediction phase.

## 4  PROPOSED AGAR METHOD

To address the limitations identified in the previous section, we now propose an improved architecture for point cloud prediction, depicted in Fig. 5. The proposed architecture preserves the state-of-art global framework composed of a DE, FP and prediction phase. However, we propose to replace current state-of-art modules, with improved versions to leverage on the point cloud semantic structure during the DE phase and to perform an adaptive combination of dynamic features in the FP phase.

### 4.1  Addressing Limitation 1: Inclusion of structural relationships between points

To overcome the lack of geometrical prior with meaningful spatial/semantic information, we propose an initial graph neural network denoted by *Spatial-Structure GNN* (SS-GNN) that processes each frame to extract for each point spatial features that carry local topological information. From the learned spatial features, we then construct a *spatio-temporal* graph that incorporates the point structural/semantic information and uses that information to build representative neighbourhoods of points. The spatio-temporal graph is processed by a proposed *graph-RNN* cells, that can extract point cloud behaviour as dynamic features. Below we present each of the proposed modules in detail.

*4.1.1  Spatial-Structure GNN (SS-GNN).* Given an input point cloud $P_t$ for each point $i$, the SS-GNN learns a spatial feature $s_{i,t}$ describing the point's local geometric structure. To learn these features SS-GNN starts by constructing a *coordinate graph* $\mathcal{G}_t^C = (P_t, \mathcal{E}_t^C)$ by taking the points $P_t$ as vertices and by building directed edges $\mathcal{E}_t^C \in \mathbf{R}^{N \times k}$ between each point to its $k$-nearest neighbours based on Euclidean distance. The SS-GNN is composed of three layers, each layer performs a graph message-passing convolution [46]. At the $h$-th layer, for a target point $i$, all its neighbouring points $j \in \mathcal{E}_i^C$ exchange a message along the edge connecting the two points. The message between points is obtained by processing the concatenation between the target point spatial feature at the previous layer $s_{i,t}^{h-1}$; the target point coordinates $p_{i,t}$; the geometry displacement between target points $i$ and it neighbours $j$ ($\Delta p_{ij}$). A symmetric function is then applied to aggregate all the messages into an updated feature for the target node. More formally, the message between two nodes ($m_{ii,t}^h$) and the output spatial features ($s_{i,t}^h$) are obtained as follows:

$$m_{ij,t}^h = \Theta_S^h(s_{i,t}^{h-1} ; p_{i,t}^l ; \Delta p_{ij}) \tag{5}$$

$$s_{i,t}^h = \bigoplus_{j \in \mathcal{E}_i^C} \left\{ m_{ij,t}^{h+1} \right\} \tag{6}$$

where $\Theta_S^h$ is a set of learnable parameters at layer $h$ abd ';' identifies the concatenation operation. The $\bigoplus$ represents an element-wise max pooling function that acts as an activation function by introducing non-linearity. It is important to note that the above operation does not involve spatio-temporal aggregation. Instead, the spatial features are learned from a single point cloud at a single timestep.

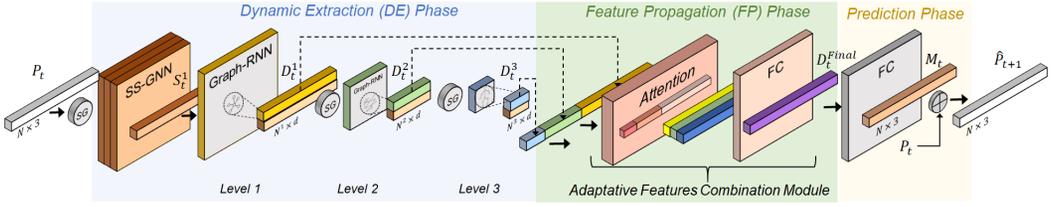

Fig. 5. **Proposed AGAR prediction architecture** composed of DE, FP and prediction phase. the DE phase, the architecture consists of an SS-GNN module followed by graph-RNN cells. The SS-GNN module extracts spatial features from the point cloud which are then utilized by the graph-RNN cells to learn dynamic features. In the FP phase, the state-of-art FP modules are replaced by a novel *Adaptative feature combination* module able to dynamically combine hierarchical features according to the scene.

### 4.1.2 Graph-RNN.
Each graph-RNN cell, at level $l$, takes as input the point coordinates, spatial and dynamic features ($P_t^l\ S_t^l\ D_t^l$) and learns updated dynamic features $D_t^{l+1}$ describing the point's dynamic behaviour. To this end, the graph-RNN cell builds a spatio-temporal graph $\mathcal{G}_t^{ST,l} = (P_{t'}^l, \mathcal{E}_t^{ST})$ between the points $P_t^l$ and $P_{t-1}^l$. Unlike the coordinate graph which is built on geometric distances, the spatio-temporal graph is built based on the spatial features distance. Specifically, for each point $i$ at time $t$, we calculate the distance between the point spatial feature $s_{i,t}$ and the spatial feature from other points in the present frame $s_{j,t}$ and in the past frame $s_{j,t-1}$. Each point $i$ is connected to its $k$-closest points in present time $t$ and its $k$-closest in past time $t-1$. By connecting points that share a common local structure, we are able to establish correspondence between points that despite not being close in the Euclidean space, they share semantic similarities and therefore they will most likely share motion vectors. Fig. 6 depicts an example of a spatio-temporal graph constructed between two frames in a fast-moving sequence of a person running (some edges are hidden for image clarity). The dashed boxes in Fig. 6 show the edges build for the points in the foot when using spatial feature distance −our approach− (upper box, in red) and the edges built if we had used coordinate distance −state-of-art approach- (lower box, in blue). The edges built on spatial feature similarity (in red) can correctly match points across time while edges based on geometry proximity would lead to incorrect grouping. As a result, the network learns dynamic features from neighbourhoods of points that share similar semantic/structural properties.

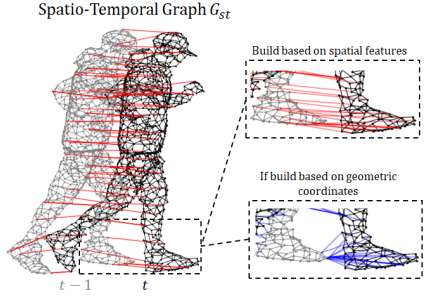

Fig. 6. Spatio-Temporal graph $G_{st}$, with some temporal edges colored in red; Dashed box depicts the difference between building the $G_{st}$ using spatial features or using point coordinates.

Similarly to the SS-GNN, the graph-RNN extracts dynamic features by performing a message-passing convolution between a point and its neighbourhoods in the spatio-temporal graph. For each target point, we learn a message for each edge by processing the concatenation of the target point dynamic feature ($d_{i,t}^l$); the neighbour point dynamic feature ($d_{j,t'}^l$) where $t'$ can be either $t$ or $t-1$; the coordinates difference ($\Delta p_{ij}$), spatial features difference ($\Delta s_{ij}$); temporal different ($\Delta t_{ij}$) between the target and neighbour point. All the messages are aggregated into a single representation to update the target point dynamic features $d_{i,t}^{l+1}$. The operation can be formalized as:

$$m_{ij,t}^l = \Theta_D^l(d_{i,t}^l;\ d_{j,t'}^l;\ \Delta p_{ij};\ \Delta s_{ij};;\Delta t_{ij}) \tag{7}$$

$$d_{i,t}^{l+1} = \bigoplus_{j \in \mathcal{E}_i^{\mathrm{ST}}} \left\{ m_{ij,t}^l \right\} \tag{8}$$

The learned spatial features are used not only to connect points with similar spatial characteristics in *both* the present and past frame but are also directly incorporated in the graph-RNN convolution. As a result, the graph-RNN learns a point dynamic behaviour taking into account structural relations to neighbourhood points. This inclusion of point spatial features in the graph-RNN cell convolution, allows the network to learn more representative dynamic features and helps to preserve the predicted point cloud shape.

## 4.2 Addressing Limitation 2: Adaptative Feature Combination

We now address the current framework limitation to generate complex motions caused by the fixed combination of dynamic features in the FP phase. To overcome the issue, we propose to replace the FP modules with an attention-based module denoted *Adaptative feature combination* represented in detail in Fig.7. Instead of using a fixed combination, the proposed module dynamically assigns an attention value to each level based on the learned features. This attention value determines the amount of influence each level will have on the predicted motion of the point.

In details, given an architecture with L hierarchical levels ($L = 3$ in the example in Fig.7 ), the proposed *Adaptative Feature combination* module takes as input the dynamic features ($D_t^1, D_t^2, ..., D_t^L$) learned in the DE phase and combines them into a single final dynamic feature ($D_t^{\mathrm{Final}}$). However, we recall that each RNN cell is preceded by a downsampling module, hence each feature needs to be up-sampled before being combined. To do this, the proposed module first interpolates the dynamic features to the same number of points as the first level and processes each independently through a refinement layer $\Theta_R^l$, to ensure the features are on a similar scale, as follows:

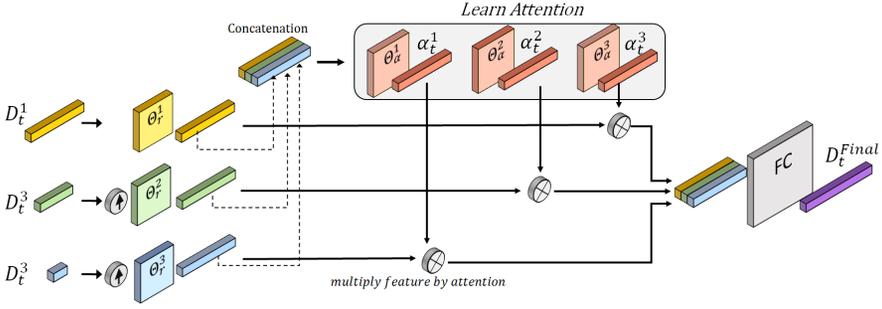

Fig. 7. **Adaptative Feature Combination Module.** Given a point cloud prediction framework with three hierarchical levels, the module takes as input dynamic features $D_t^1, D_t^2, D_t^3$ and outputs a single final dynamic feature $D^{Final}$

$$\psi(d_{i,t}^l) = \sigma\left(\Theta_R^l \{d_{i,t}^l\}\right) \tag{9}$$

where $d_{i,t}^l$ are the interpolated features to original number of points, $\psi(d_{i,t}^l)$ are the outputted refined features and $\sigma$ is the activation function. To learn scalar attention values $\alpha_{i,t}^l$, the network concatenates the refined features from all levels and processes them through learnable parameters $\Theta_\alpha^l$ as follows:

$$\alpha_{i,t}^l = \sigma\left(\Theta_\alpha^l \{\psi(d_{i,t}^1); \psi(d_{i,t}^2); \psi(d_{i,t}^3)\}\right) \tag{10}$$

The refined dynamic features $\psi(d_{i,t}^l)$ are then multiplied by their respective attention value. Hence, the $\alpha$ value reflects the *influence* that the learned feature has on the predicted motion, allowing the network to adjust the contribution of each level to the predicted motion. Namely,

$$\Psi(d_{i,t}^l) = \psi(d_{i,t}^l) \times \alpha_{i,t}^l \tag{11}$$

Lastly, the dynamic features post-attention module $\Psi(d_{i,t}^l)$ are combined by a single learnable layer ($\Theta_{FC}$) into the final dynamic features $d_{i,t}^{\text{Final}} \in D_t^{\text{Final}}$.

$$d_{i,t}^{Final} = \sigma\left(\Theta_{FC}\{\Psi(d_{i,t}^1); \Psi(d_{i,t}^2); \Psi(d_{i,t}^3)\}\right) \tag{12}$$

*4.2.1 Explainablity of the Adaptative feature combination module.* A key benefit of the *Adaptative feature combination* module is that its underlying mechanism can be visualized and explained. This can be seen in Fig. 8, which illustrates how the proposed module combines dynamic features to produce motion vectors given two point cloud sequences (*Man-Running and Woman-Dancing*). For each sequence, Fig. 8 depicts: the PCA of the dynamic features learned at the DE phase; the learned attention values per point; the individual motion vectors[1] produced at each level in the proposed *Adaptative* architecture and in the *Classic-FP* architecture (previously presented in Section 2.2 and Fig.2).

In the *Man-Running* sequence depicted in Fig. 8 (a), at the first level the network assigns high attention values ($\alpha_t^1$) to the arms and low attention values to the points in the rest of the body. As a result, the predicted motion of the points in the arms is heavily influenced by local motions, while in the rest of the body, the local motions have a very small influence on prediction. The network exhibits similar selective behaviour at the second level, assigning higher attention to the points in the left foot, increasing the influence of the dynamic features $D_t^2$ have on the motion of the foot. In the third and final level, the network learned non-zero attention values $\alpha_t^3$ for the majority of the body. As a result, in the *Man-Running* sequence, the global motion is the primary contributor to the predicted motion of the points, with the exception of the arm and the foot regions where the prediction is given by a motion combination from multiple levels.

---

[1]For the sake of image clarity the motion vectors were uniformly sampled

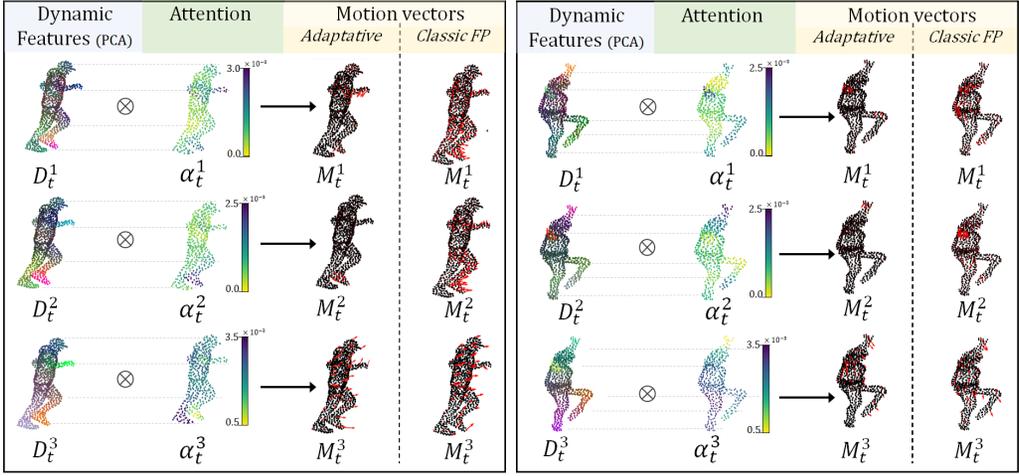

(a) *Man-Running*.                                   (b) *Woman-Dancing*.

Fig. 8. *Adaptative Features Combination* operation. Example of how the proposed module adaptively combines local and global motion for different points, and comparison with motion obtained with *Classic-FP*.

Similar considerations can be derived from the second example *Woman-Dancing*, in which the learned global motions are an accurate descriptor for the majority of the points, except for certain regions with more local movements. The *Adaptative feature combination* module is able to distinguish between regions and combine properly the different levels of motions based on the distinction. It is worth noting that different attention values are learned for the *Man-Running* and *Woman-Dancing* sequences, demonstrating the network's ability to adapt the attention according to the characteristics of the input data. In summary, the proposed *Adaptative feature combination* module combines features in an adaptive manner, allowing it to **control the composition of global and local motions** that best describes the motion of each point. This adaptive operation can be understood and explained through visualization, which may be beneficial for future research on developing more expressive architectures.

## 5  IMPLEMENTATION

In this section, we describe the datasets and implementation detail of our proposed method. To ensure reproducibility, we release the dataset and the source code of our architecture as well as the benchmarking methods[2].

---

[2]https://github.com/pedro-dm-gomes/AGAR

## 5.1 Datasets

In our experiments, we considered the following datasets:

**Mixamo Human Bodies Activities**: Synthetically human motions generated following [34], using the online service Mixamo [14] and Blender software [1]. Despite being synthetic the dataset provides an accurate representation of real-world movements. We create 152 test sequences and 9,375 training sequences (further augmented by randomly changing movement direction, speed, and the body starting position during training). Each training sequence consists of approximately 50 frames, and for each sequence, we sample $T = 12$ consecutive frames as inputs to the model during training. Similarly, the testing sequences are composed of 12 frames. Each frame in the dataset contains a point cloud consisting of 1,000 points, which we found to be sufficient for capturing a rich and detailed representation of the human body.

**CWIPC-SXR Human motions** [30]: Real-world human motions in social settings. The dataset consists of 21 dynamic sequences. For each sequence, we sampled the first 60 frames from a capture rate of 30 fps to 10 fps, resulting in 21 sequences of 15 frames each ($T = 15$). Given its reduced size, this dataset is not used for training but only for testing. To ensure consistency with the training data (*Mixamo*), we downsampled the dataset to 1,000 points per frame before feeding it into the model.

**JPEG Pleno Voxelized Full Bodies** [5]: Real-world human bodies. The dataset is composed of four sequences known as longdress, loot, redandblack, and soldier. This dataset is not used for training, only for testing. Each sequence is downsampled to 12 frames ($T = 12$) and 1,000 points.

**Moving MNIST Point Cloud**: Created by converting the MNIST dataset of handwritten digits into moving point cloud, as previous works [6]. The sequences are generated by applying rigid motion at random to each digit. Each sequence contains 20 frames ($T = 20$) with either 128 (1 digit) or 256 points (2 digits).

**Argoverse** [3]: Large scale automotive dataset. We use the same train and test data as in PointRNN [6]. The dataset contains 910 training sequences and 209 test sequences. Each sequence contains 20 frames ($T = 20$) and each frame is downsampled to 1024 points.

**MSRAction3D** [17]: Real-world human motion performing annotated actions. The dataset consists of 567 Kinect depth videos, with 20 action categories. We sampled each point cloud to 1024 points, and use the same training and test conditions as works [8, 19].

## 5.2 Benchmarking

This subsection outlines the tasks, as well as the state-of-the-art benchmark methods used for comparison.

*5.2.1 Prediction Task.* In the prediction task, we consider both short-term and long-term predictions. In short-term prediction, at each iteration, the network takes as input the ground truth frame $P_t$ to predict the next frame $\hat{P}_{t+1}$. At the following prediction step, the network will be predicting $\hat{P}_{t+2}$ having as input the ground truth $P_{t+1}$. This is repeated till the end of the sequence. In long-term prediction, the predicted frame from the previous interaction $\hat{P}_t$ is used as input to predict the next frame $\hat{P}_{t+1}$. In long-term prediction only the later half ($T/2$) of sequence is predicted using this strategy. For benchmarking, since point cloud prediction of human bodies is a mostly an unexplored topic, the range of possible choices of baseline methods to compare our work is limited. Moreover, many of the existing point-based RNN point cloud prediction methods designed for automobile scenes do not provide the necessary materials to be replicated. Therefore, besides selecting the most related works available, we adapted several methods that, while not originally designed for point cloud prediction, are well-recognized in the field of point cloud sequence processing. For the point cloud prediction task, we consider the following as baseline models: (1) *Copy-Last-input* model which simply copies the past point cloud frame instead of predicting it; (2) PointPWC-Net [45] a hierarchical point-based architecture to extract the motion flow between two frames; (3) FlowStep3D [15] a hierarchical point-based architecture to extract learned motion flow between two frames via RNN cells; (4) PSTNet [8] a hierarchical point-based architecture for action classification of human body sequences; (5) PointRNN [6] ($k$-NN): point-based RNN architecture presented in Section 2.2. Both PointPWC-Net and FlowStep3D were originally designed to learn the motion flow between two frames. To extend these two models to the task of predicting future frames, we incorporate a prediction phase into their architectures. This prediction phase refines the extracted motion flow via fully connected layers and calculates a predicted point cloud at the next time step. Similarly, the PST-Net architecture, designed for action classification, is adapted for the prediction task by adding an FP phase (with *Classic-FP*) to propagate the learned features to the original number of points, followed by a prediction phase to generate a prediction of the point cloud at the next timestep given the propagated features. To differentiate the adapted models from their original counterparts, we denote the adapted models for the prediction task as PointPWC-Net-*pred*, FlowStep3D-*pred* and PST-Net-*pred* respectively.

*5.2.2 Action Classification Task.* To study the generalizability of the proposed AGAR framework for dynamic feature learning, we extended its application to the classification task. In this task, the AGAR takes a point cloud sequence as input and outputs a classification score. To adapt AGAR for the classification task, we discarded the FP phase and the prediction phase. Instead, the dynamic features from the last level are max-pooled to form a global feature, which is used to generate the classification score. We denoted this architecture adapted for classification tasks as AGAR-*cls*. Given human action classification from point clouds sequences a well-studied problem we compare AGAR-*cls* to well-established methods such as MeteorNet [19], PSTNet [8], P4Transformer [7] without adaptations.

## 5.3 AGAR Architecture details

For the prediction and classification tasks, we implemented AGAR and an AGAR-*cls* (adapted for classification) architectures with three hierarchical levels ($L = 3$) respectively. In both cases, the SS-GNN in the first level consists of three layers with 64, 128, and 128 dimensions respectively. Each level contains a graph-RNN cell that learns dynamic features with 128 dimensions. The number of nearest neighbours ($k$) is 8 for all graph-RNN cells. Between each graph-RNN cell, the point cloud is downsampled by a factor of 4. All the models are trained using the Adam optimizer, with a learning rate of $10^{-4}$ for $500,000$ interactions. In the training phase, we utilize a batch size of 16 for the Mixamo Human Bodies dataset, 32 for the MNIST dataset, 4 for the Argoverse dataset, and 32 for MSRAction3D. For all models, the gradients are clipped in the range [5, 5].

## 5.4 Training and Metrics

The AGAR architecture has multiple end-to-end parameters, trained in a self-supervised fashion by comparing the predicted point cloud $\hat{P}_{t+1}$ with the target point cloud $P_{t+1}$. Unlike supervised methods [15, 19, 40, 45], which require the ground-truth motion flow to train the network, in a self-supervised setting the ground-truth data can be obtained from the input data itself. This technique allows us to train on a dataset of deformable dynamic point clouds, such as human bodies dataset [5, 14, 30], where annotated ground-truth motion vectors are not available.

*5.4.1 Training Metrics.* To measure the difference between the predicted point cloud and the ground-truth point cloud during training, we employ the commonly used chamfer distance (CD) [13] and earth's moving distance (EMD) [2]. These metrics are defined as the following:

*Chamfer distance (CD) :* The CD measures the distance between each point in the predicted point cloud and its closest target point in the reference point cloud, and vice-versa.

$$d_{CD}(P, \hat{P}) = \frac{1}{n} \sum_{p \in P} \min_{\hat{p} \in \hat{P}} ||p - \hat{p}||^2 + \frac{1}{n} \sum_{\hat{p} \in \hat{P}} \min_{p \in P} ||\hat{p} - p||^2 \qquad (13)$$

*Earth's moving distance (EMD):* The EMD solves an optimization problem, by finding the optimal point-wise bijection mapping between two point clouds $\theta : P \rightarrow \hat{P}$. The EMD distance is then given by the distance of the points at both ends of this mapping, as follows:

$$d_{EMD}(P, \hat{P}) = \min_{\theta:P \rightarrow \hat{P}} \sum_{p \in P} ||p - \theta(p)||^2. \qquad (14)$$

Although the EMD and CD metrics are commonly used in point cloud analysis, they may not always provide an accurate measure of similarity. The CD only considers the nearest neighbour of a point and does not take into account the global distribution of points. On the other hand, EMD tries to find a unique mapping between two point clouds. However, in most cases a unique mapping is realistically impossible, resulting in a measurement that is rarely correct for all points. Since CD and EMD measure different notions of similarity with different shortcomings, we use a combination of both metrics as the loss function in order to make the loss function more robust, as follows:

$$\mathcal{L}(P, \hat{P}) = d_{CD}(P, \hat{P}) + d_{EMD}(P, \hat{P}) \qquad (15)$$

*5.4.2 Evaluation Metrics.* To evaluate our model we used the CD and EMD metrics also used for training. However since CD and EMD measure the similarity between two point clouds by averaging the distance across all points, they tend to flatten their distance scores towards zero values. This is because in a point cloud, the majority of points are perfectly predicted (either no motion or little motion), and most of the high prediction errors are concentrated in small areas of high or complex motion. Therefore to better evaluate the model's ability to predict complex motions, besides the CD and EMD we also consider the following additional evaluation metric, defined as:

*Chamfer distance of the top %5 worst points (CD Top %5):* This metric returns the average CD distance of the 5% of points with the worst predictions (i.e., points with the farthest distance to their closest point). We found that this CD Top %5 focuses on the regions where the body performs complex motions and provides the best correlation with the visual quality. To the best of our knowledge, we are the first to work to present results using CD top 5% metric.

## 6 EXPERIMENTAL RESULTS

In this section, we present and discuss the results of our proposed AGAR method, described in Section 4 for each task and dataset. We begin by presenting and discussing the results point cloud prediction of human body motions, which is the main goal of this paper. Next, we present the experimental results for the prediction of rigid point clouds (i.e., moving digits and automobile scenes). This is followed by the results for action classification on human body motions. Lastly, we present an ablation study on the prediction of human body motions.

### 6.1 Prediction of Synthetic Human Bodies Motions - Mixamo human bodies

The short-term prediction results from *Mixamo* dataset of human body activities can be found in Table 2 and Fig. 9 depicts prediction examples for two sequences. In addition to evaluating the AGAR architecture with *Adaptive feature combination* described in Section 4, we also evaluate a modified AGAR architecture where the *Adaptive feature combination* is replaced by *Classic-FP*. The results in Table 2 show PointRNN and both variations of the AGAR architecture outperformed the remaining methods by a large margin, demonstrating the superiority of the RNN architecture for interactive prediction. Furthermore, both AGAR architectures consistently outperform PointRNN, achieving lower prediction error in all three metrics (CD, EMD, CD Top%5). Notably, the AGAR with *Adaptive feature combination*) achieves an EMD error of 58.2, surpassing PointRNN's EMD error of 68.0 with a 10.2 gain. This gain is especially significant for deformable objects since shape distortion has a high visual impact. This is particularly noticeable in the last frame ($t = 10$) of the "Woman-Turning" sequence (in Fig. 9), where the AGAR prediction suffers less deformation compared to the PointRNN prediction. In the following, we analyze the improvement provided by each component of the proposed AGAR method to better understand the impact of each limitation on the prediction task.

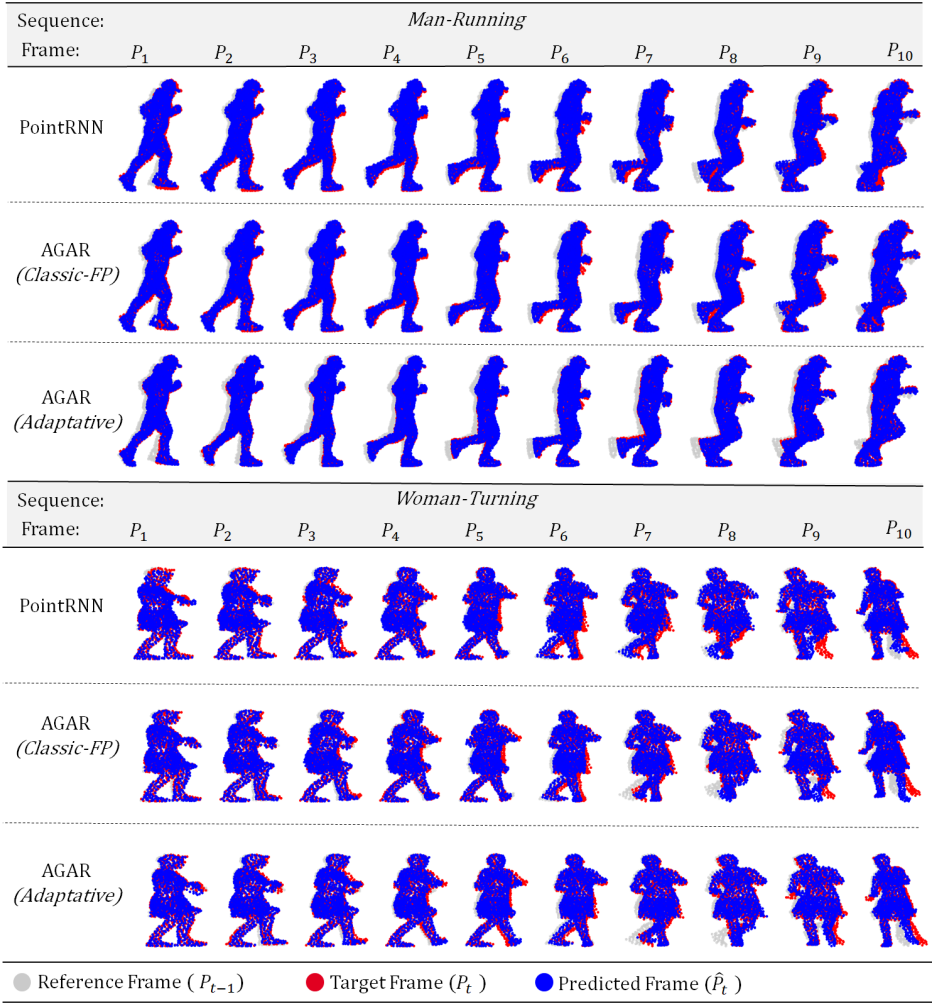

Fig. 9. Example of prediction of human bodies activities on the Mixamo dataset.

| Mixamo (Synthetic Human bodies dataset) | | | |
|---|---|---|---|
| Model | CD | EMD | CD Top 5% |
| Copy-Last-input | 0.1056 | 123.4 | 0.2691 |
| PointPWC-Net-*pred* [45] | 0.09358 | 118.5 | 0.2601 |
| FlowStep3D-*pred* [15] | 0.09153 | 115.6 | 0.2575 |
| PSTNet-*pred* [8] | 0.08984 | 114.1 | 0.2556 |
| PointRNN [6] | 0.00351 | 68.0 | 0.1593 |
| AGAR    *Classic-FP* | 0.00262 | 59.6 | 0.1412 |
| AGAR    *Adaptative* | **0.00254** | **58.2** | **0.1346** |

Table 2. Point cloud prediction results on the Mixamo dataset

| Mixamo (Synthetic human bodies dataset) | | | | | |
|---|---|---|---|---|---|
| Model | Type of graph | Spatial features | CD | EMD | CD Top 5% |
| AGAR (*Classic-FP*)    (i) spatio-temporal | ✓ | | **0.00262** | **59.6** | **0.1410** |
| AGAR (*Classic-FP*)    (i) spatio-temporal | ✗ | | 0.00341 | 67.0 | 0.1602 |
| AGAR (*Classic-FP*)    (ii) only temporal | ✓ | | 0.00266 | 60.0 | 0.1417 |

Table 3. Comparison of three variations of the AGAR framework demonstrating gain from the including structural relations between points in the spatio-temporal graph.

To understand the impact of combining features in an adaptative manner we compare the AGAR with *Adaptive feature combination* and the AGAR with *Classic-FP*. Table 2 shows the AGAR with *Adaptive feature combination* achieves a lower prediction error compared to the AGAR with *Classic-FP*. While the error improvement in terms of CD and EMD is relatively small, the CD Top 5% metric, which is more sensitive to local distortion, shows a clear improvement in the AGAR with *Adaptive feature combination*. The superior performance of adaptively combining dynamic features can also be seen by looking at the visual results in Fig.9. We can notice the AGAR with *Adaptive features combination* predicts better specific regions such as the hands and the legs, which involve complex motions. This improvement is due to the module's ability to generate refined motion predictions required in these regions. **These results show the clear advantage of adaptively combining dynamic features to predict complex motions.**

To understand the advantages of incorporating the structural relations between points when dynamic learning features, in Table 3 we compare: i) an AGAR architecture; ii) an AGAR model that does not learn spatial features (without the SS-GNN module). Hence does not take the structural relation between the point into account, when learning dynamic features; iii) an AGAR model that learns spatial features, but builds only a temporal graph i.e., a *k-nn* graph is built only connecting each point of the frame $t$ with points in frame $t-1$ (the total number of neighbours $k = 8$ remains the same for fairness). All three model variations have a *Classic-FP* phase. The results show there is a relatively small gain in building a complete spatio-temporal graph but significant improvement by learning spatial features. It is worth noticing, that the *CD Top 5%* (the most sensitive metric to point cloud local shape distortion) is significantly lower in the model that learns spatial features compared to the model that does not learn spatial features. This demonstrates that while both models are able to capture the overall motion, **the inclusion of spatial features in the DE phase significantly improves the accuracy and preservation of the predicted point cloud's shape.**

| JPEG and CWIPC-SXR Real-world human bodies dataset | | | | | | |
|---|---|---|---|---|---|---|
| Method | JPEG | | | CWIPC-SXR | | |
| | CD | EMD | CD Top 5% | CD | EMD | CD Top 5% |
| Copy Last Input | 0.00118 | 42.0 | 0.09001 | 0.00295 | 43.2 | 0.12915 |
| PointRNN | 0.00109 | 41.3 | 0.083461 | 0.00157 | 43.4 | 0.10973 |
| AGAR *Classic-FP* | 0.00101 | 38.6 | 0.08172 | **0.00150** | 40.8 | **0.10655** |
| *Adaptative* | **0.00095** | **37.4** | **0.07754** | 0.00155 | **39.8** | 0.10760 |

Table 4. Prediction Error for real-world human bodies dataset.

| MNIST Dataset | | | | |
|---|---|---|---|---|
| Method | Long-Term prediction | | | |
| | 1 digit | | 2 digits | |
| | CD | EMD | CD | EMD |
| Copy Last Input | 262.46 | 15.94 | 140.14 | 15.8 |
| PointRNN | 2.25 | 2.52 | 14.54 | 6.42 |
| AGAR *Classic-FP* | **0.88** | **1.52** | **1.67** | **2.60** |
| *Adaptative* | 0.96 | 1.60 | 1.75 | 2.62 |

Table 5. Prediction error on the MNIST dataset.

## 6.2 Prediction of Real Human Bodies Motions - JPEG and CWIPC-SXR dataset

We now turn our focus to real-world human bodies datasets: the JPEG and CWIPC-SXR datasets. Since both the JPEG and CWIPC-SXR datasets are too small to train models, they are only used for the evaluation of the models trained on the Mixamo dataset. Table 4 depicts the short-term prediction results from real-world data from the JPEG dataset, and the CWIPC-SXR dataset. It can be noted, the *Copy-last-input* has significantly lower prediction error in real-world datasets compared to the error on the Mixamo dataset. In the JPEG and CWIPC-SXR dataset, the point clouds were acquired from real test subjects only allowed to move in a small area, resulting in a lower magnitude of motion compared to the Mixamo dataset. Despite the lack of motion, the AGAR model is able to make accurate predictions and achieved the smallest prediction error across all metrics. The small improvement of the *Adaptive* combination over the *Classic-FP* can be attributed to the low magnitude of motion in the dataset. Importantly these results demonstrate that the AGAR model trained on synthetic human motions datasets can be effectively applied to real-world human motions datasets despite the large disparity in motion magnitudes between the two datasets.

## 6.3 Prediction of Rigid Object - MNIST dataset Moving Digits

The simplicity of representation and movements performed by the MNIST dataset makes it the ideal dataset to test the long-term prediction of the proposed AGAR method. Long-term prediction is when the network uses its output predictions at a time-step as input for the subsequent time-step. We present the prediction results for the MNIST dataset in Table 5, and prediction examples in Fig.10. Table 5 shows the AGAR model has superior prediction performance compared to the PointRNN model. This performance gap is particularly large for point clouds containing two digits. In two-digits, the Point-RNN CD prediction error is 14.54 whereas the AGAR (*Classic-FP*) CD error is 1.67. This large gain is due to the AGAR's ability to learn spatial features, which allows it to understand the structure to discern the two distinct shapes. This improvement can be seen in Figure 10, where all the evaluated models exhibit a progressive loss of shape, however, the AGAR model suffers from significantly less deformation compared to Point-RNN. This visualization demonstrates that the AGAR is better at preserving the spatial structure over time, a direct effect of learning the point cloud spatial structure. Lastly, it can be noted the AGAR model with *Adaptive feature combination* and with the *Classic-FP* have a similar prediction error, also seen in the example in Fig. 10. The reason being in the moving digits dataset there are no complex motions (i.e, the digits perform simple rigid translation), as such the control over the motion provided by the *Adaptative features combination* module is just unnecessary parameterization and does not translate into more accurate predictions.

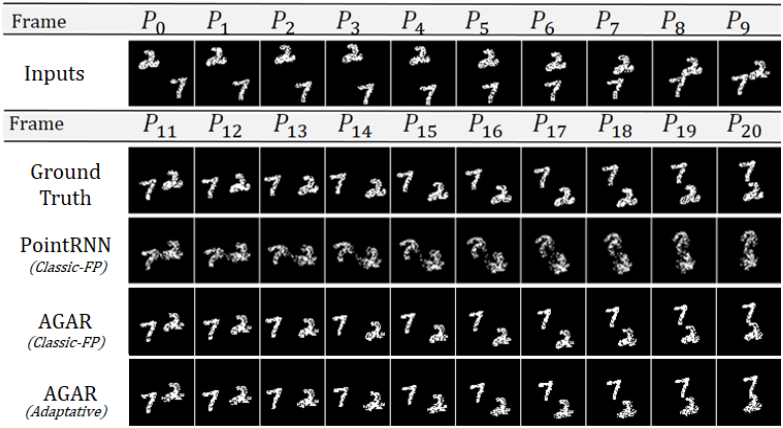

Fig. 10. Long-term predictions examples of MNIST sequences.

| Argoverse (Automobile scenes dataset) | | |
|---|---|---|
| Method | Long-Term Prediction | |
| | CD | EMD |
| Copy Last Input | 0.5812 | 1092.3 |
| PointRNN | **0.2541** | 895.28 |
| AGAR *Classic-FP* | 0.2680 | **875.22** |
| AGAR *Adaptative* | 0.2839 | 893.24 |

Table 6. Prediction error for the Argoverse dataset.

## 6.4 Prediction of Automobile scenes- Argoverse dataset

Table 6 shows the results of training and evaluating the AGAR model and the PointRNN baseline with the Argoverse automobile dataset. Not surprisingly, both methods achieved similar prediction errors. This was an expected result, as the characteristics of deformable bodies on which AGAR relies are not present in the automobile dataset. More specifically the structural information in the data is not informative and reliable enough for the SS-GNN module to leverage when learning features. Similarly, the data does not perform complex motions that would require *Adaptive features combination* module. Hence, the inclusion of both modules is not translated into a meaningful gain. However, despite being designed for deformable objects, the results demonstrate that the proposed

minimal

| MSR Action | | | | | | | | | |
|---|---|---|---|---|---|---|---|---|---|
| Method | Input | Accuracy | | | | | | | |
| | | #Frames | | | | | | | |
| | | 1 | 4 | 8 | 12 | 16 | 18 | 20 | 24 |
| Vieira *et al.* [33] | depth | | | | | | | 78.20 | |
| Klaser *et al.* [16] | depth | | | | | 81.43 | | | |
| PointNet++ [29] | point | 61.61 | | | | | | | |
| MeteorNet [19] | point | | 78.11 | 81.14 | 86.53 | 88.21 | | | 88.50 |
| PSTNet [8] | point | | 81.14 | 83.50 | 87.88 | **89.90** | | | **91.20** |
| P4Transformer [7] | point | | 80.13 | 83.17 | 87.54 | 89.56 | | | 90.94 |
| AGAR-*cls* | point | | **81.48** | **87.20** | **88.21** | 88.55 | | | 90.09 |

Table 7. Action recognition accuracy (%) on the MSR-Action3D dataset for 4, 8, 12, 16, 24 frames as input.

| Mixamo (Synthetic human bodies dataset) | | | |
|---|---|---|---|
| Number of levels | CD | EMD | CD Top 5% |
| 1 | 0.00296 | 65.4 | 0.166 |
| 2 | 0.00276 | 61.2 | 0.1461 |
| 3 | **0.00262** | 59.6 | **0.1412** |
| 4 | 0.00290 | 62.0 | 0.14745 |

Table 8. Effect of the number of levels on the AGAR framework.

| Mixamo (Synthetic human bodies dataset) | | | |
|---|---|---|---|
| Number of neigborhoods | CD | EMD | CD Top 5% |
| 4 | 0.00290 | 62.8 | 0.1489 |
| 8 | 0.00262 | 59.6 | 0.1412 |
| 12 | **0.00264** | **58.9** | **0.1407** |

Table 9. Effect of the number of neighbours on the AGAR framework.

| Mixamo (Synthetic human bodies dataset) | | | |
|---|---|---|---|
| Down-sampling factor | CD | EMD | CD Top 5% |
| 1 | 0.00314 | 65.0 | 0.160 |
| 2 | **0.00259** | **58.7** | **0.138** |
| 4 | 0.00262 | 59.6 | 0.1412 |

Table 10. Effect of the downsampling factor on the AGAR framework.

AGAR is still capable to process and capturing the overall correct movement from point clouds of automobile scenes.

## 6.5 Action Recognition of Human Motions - MSR3DAction Dataset

Table 7 presents the results of the action recognition task on the MSRAction dataset. As described in Section 5.2, here we compare the AGAR-*cls* with multiple well-known methodologies optimized for action classification. In the table, we provide the accuracy of different methods given input point cloud sequences of 4, 8, 12, 16, 24 frames. When looking at a shorter sequence (less than 12 frames), the proposed AGAR-*cls* outperformed state-of-art methods. Notably, for sequences of 8 frames, the AGAR-*cls* achieved 87.2% accuracy a 5% improvement over the PSTNet and P4Transfomer. However such gain is lost, for sequences longer than 12 frames, where both PSTNet and P4Transformer are slightly better than AGAR-*cls*. The reason for this decline in performance can be attributed to the RNN architecture of the AGAR-*cls* framework. Accurate action recognition requires the model to retain information about early movements throughout the entire sequence. In the AGAR-*cls* this information is retained in RNN hidden states. However, these states are continuously updated each iteration, as a result, the older information is not efficiently retained as PSTNet which processes all frames simultaneously. Despite this limitation, the results demonstrate the AGAR-*cls* ability to capture complex motions from human body point clouds, making it a promising model for action recognition tasks, especially for shorter sequences. Furthermore, the understanding of the dynamic feature's role in capturing complex human motions presented in Section 3, can also provide valuable insight for action recognition. The understanding of how the composition of local and global motions leads to a class prediction can help in understanding why certain actions are misclassified, leading to the design of more accurate architectures.

## 6.6 Ablation Study

To gain a deeper understanding of our proposed architecture, an ablation study is conducted on the Mixamo Synthetic dataset for short-term prediction. Table 8, Table 9 and Table 10 show how each parameter influences the performance of the network.

**The number of levels** (Table 8): The best results were achieved with architecture with three hierarchical levels ($L = 3$), showing that increasing the number of levels does not necessarily lead to superior performance. However, a minimum number of levels does impact positively the accuracy, confirming the importance of hierarchical learning.

**Neighborhood size** (Table 9): The results show an increasing number of neighbours points ($k$) improves the model performance. However, increasing neighbours also significantly increases the memory required to train the model. This illustrates one of the main limitations of the current deep learning frameworks, which is the high GPU memory requirements. This limitation was not addressed in this paper.

**The downsampling factor** (Table 10): Given a point cloud with $1,000$ points, a down-sample by a factor of 2 for each level leads to the best results. (i.e, 500, 250, and 125 points at levels 1, 2 and 3 respectively). Using a downsampling factor of 1 (i.e, no sampling between layers) resulted in the worst performance, which was similar to the performance obtained using a single level ($L = 1$). This demonstrates that the improvement gained from using hierarchical architecture levels is due to learning features from neighbourhoods at different scales.

## 7 CONCLUSION

The goal of this paper is to improve current prediction frameworks for point clouds representing deformable 3D objects, with a focus on human bodies motions. To reach this goal, we investigated the current state-of-the-art point-based RNN prediction framework and identified its limitations when processing deformable shapes and complex motions present in deformable objects. To overcome these limitations, we propose an improved architecture for dynamic point cloud processing. This architecture includes an initial graph-based module that learns the structural relations of point clouds as spatial features. From the spatial features, we then construct spatio-temporal graphs. This module is followed by a hierarchy of graph-RNN cells, to extract dynamics features from the spatio-temporal graphs taking the learned structural relations between points into account. Lastly, as a key novelty, we propose a module able to combine dynamic features learned by the graph-RNN cells in a *adaptative* manner Our proposed module assigns a level of attention to each hierarchical feature in order to control the composition of local and global motion that best describes each point motion. Notably, the adaptive combination module inner-working can be visualized and understood, opening the door to future research to gain insights to develop more expressive architectures. Our experimental results demonstrate the superiority of the proposed architecture in motion prediction and in action classification of deformable objects. We also showed this improvement is due to the method's ability to exploit the spatial structure of the point cloud to extract more representative dynamic features, as well as the adaptive combination of the dynamic features to predict complex motions.

## ACKNOWLEDGMENTS

This work has been partially funded by CISCO under the Academic Donation Scheme and by the EPSRF-SFI grant EP/T03324X/1

## REFERENCES

[1] Blender Foundation. 1994. Blender - a 3D modelling and rendering package. Retrieved June 29, 2022 from http://www.blender.org

[2] Sergio Cabello, Panos Giannopoulos, Christian Knauer, and Günter Rote. 2008. Matching Point Sets with Respect to the Earth Mover's Distance. *Computational Geometry* (2008).

[3] Ming-Fang Chang, John Lambert, Patsorn Sangkloy, Jagjeet Singh, Slawomir Bak, Andrew Hartnett, De Wang, Peter Carr, Simon Lucey, Deva Ramanan, et al. 2019. Argoverse: 3d Tracking and Forecasting with Rich Maps. In *Proceedings of the IEEE Conference on Computer Vision and Pattern Recognition*.

[4] Girum G. Demisse, Djamila Aouada, and Björn Ottersten. 2018. Deformation-Based 3D Facial Expression Representation. *ACM Transactions on Multimedia Computing, Communications, and Applications* (2018).

[5] Eugene d'Eon, Bob Harrison, Taos Myers, and Philip A. Chou. 2017. 8i Voxelized Full Bodies A Voxelized Point Cloud Dataset. In *ISO/IEC JTC1/SC29 Joint WG11/WG1 (MPEG/JPEG) input document WG11M40059/WG1M74006*.

[6] Hehe Fan and Yi Yang. 2019. PointRNN: Point Recurrent Neural Network for Moving Point Cloud Processing. *arXiv preprint arXiv:1910.08287* (2019).

[7] Hehe Fan, Yi Yang, and Mohan Kankanhalli. 2022. Point Spatio-temporal Transformer Networks for Point Cloud Video Modeling. *IEEE Transactions on Pattern Analysis and Machine Intelligence* (2022).

[8] Hehe Fan, Xin Yu, Yuhang Ding, Yi Yang, and Mohan Kankanhalli. 2020. PSTNet: Point Spatio-temporal Convolution on Point Cloud Sequences. In *Proceedings of the International Conference on Learning Representations*.

[9] Yifan Feng, Zizhao Zhang, Xibin Zhao, Rongrong Ji, and Yue Gao. 2018. Gvcnn: Group-view Convolutional Neural Networks for 3d Shape Recognition. In *Proceedings of the IEEE Conference on Computer Vision and Pattern Recognition*.

[10] Pedro Gomes. 2021. Graph-Based Network for Dynamic Point Cloud Prediction. In *Proceedings of the 12th ACM Multimedia Systems Conference*.

[11] Pedro Gomes, Silvia Rossi, and Laura Toni. 2021. Spatio-temporal Graph-RNN for Point Cloud Prediction. In *Proceedings of the International Conference on Image Processing*.

[12] Pedro Gomes, Silvia Rossi, and Laura Toni. 2022. Explaining Hierarchical Features in Dynamic Point Cloud Processing. In *Proceedings of IEEE Picture Coding Symposium*.

[13] Tianxin Huang and Yong Liu. 2019. 3D Point Cloud Geometry Compression on Deep Learning. In *Proceedings of the 27th ACM International Conference on Multimedia*.

[14] Adobe Inc. 2008. Mixamo: Animated 3D characters for games, film, and more. Retrieved June 29, 2022 from https://www.mixamo.com/

[15] Yair Kittenplon, Yonina C Eldar, and Dan Raviv. 2021. Flowstep3D: Model Unrolling for Self-Supervised Scene Flow Estimation. In *Proceedings of the IEEE Conference on Computer Vision and Pattern Recognition*.

[16] Alexander Klaser, Marcin Marszałek, and Cordelia Schmid. 2008. A Spatio-temporal Descriptor based on 3d-gradients. In *Proceedings of the British Machine Vision Conference*.

[17] Wanqing Li, Zhengyou Zhang, and Zicheng Liu. 2010. Action Recognition Based on a Bag of 3d Points. In *Proceedings of the IEEE Conference on Computer Vision and Pattern Recognition*.

[18] Xingyu Liu, Charles R. Qi, and Leonidas J. Guibas. 2019. Flownet3d: Learning Scene Flow in 3d Point Clouds. In *Proceedings of the IEEE Conference on Computer Vision and Pattern Recognition*.

[19] Xingyu Liu, Mengyuan Yan, and Jeannette Bohg. 2019. MeteorNet: Deep Learning on Dynamic 3d Point Cloud Sequences. In *Proceedings of the IEEE International Conference on Computer Vision*.

[20] Chris Xiaoxuan Lu, Muhamad Risqi U. Saputra, Peijun Zhao, Yasin Almalioglu, Pedro P. B. de Gusmao, Changhao Chen, Ke Sun, Niki Trigoni, and Andrew Markham. 2020. MilliEgo: Single-Chip MmWave Radar Aided Egomotion Estimation via Deep Sensor Fusion. In *Proceedings of the 18th Conference on Embedded Networked Sensor Systems*.

[21] Fan Lu, Guang Chen, Zhijun Li, Lijun Zhang, Yinlong Liu, Sanqing Qu, and Alois Knoll. 2021. MoNet: Motion-based Point Cloud Prediction Network. *IEEE Transactions on Intelligent Transportation Systems* (2021).

[22] Daniel Maturana and Sebastian Scherer. 2015. Voxnet: A 3d Convolutional Neural Network for Real-time Object Recognition. In *Proceedings of the IEEE International Conference on Intelligent Robots and Systems*.

[23] Moritz Menze and Andreas Geiger. 2015. Object Scene Flow for Autonomous Vehicles. In *Proceedings of the IEEE Conference on Computer Vision and Pattern Recognition*.

[24] Yuecong Min, Yanxiao Zhang, Xiujuan Chai, and Xilin Chen. 2020. An Efficient Pointlstm for Point Clouds Based Gesture Recognition. In *Proceedings of the IEEE Conference on Computer Vision and Pattern Recognition*.

[25] Himangi Mittal, Brian Okorn, and David Held. 2020. Just Go With the Flow: Self-supervised Scene Flow Estimation. In *Proceedings of the IEEE Conference on Computer Vision and Pattern Recognition*.

[26] Liangliang Nan, Andrei Sharf, Hao Zhang, Daniel Cohen-Or, and Baoquan Chen. 2010. SmartBoxes for Interactive Urban Reconstruction. *ACM Transactions on Graphics* (2010).

[27] Weizhi Nie, Qi Liang, Yixin Wang, Xing Wei, and Yuting Su. 2020. MMFN: Multimodal Information Fusion Networks for 3D Model Classification and Retrieval. *ACM Transactions on Multimedia Computing, Communications, and Applications* (2020).

[28] Charles R. Qi, Hao Su, Kaichun Mo, and Leonidas J. Guibas. 2017. Pointnet: Deep Learning on Point Sets for 3d Classification and Segmentation. In *Proceedings of the IEEE Conference on Computer Vision and Pattern Recognition*.

[29] Charles Ruizhongtai Qi, Li Yi, Hao Su, and Leonidas J Guibas. 2017. Pointnet++: Deep Hierarchical Feature Learning on Point Sets in a Metric Space. In *Proceedings of the Conference on Neural Information Processing Systems*.

[30] Ignacio Reimat, Evangelos Alexiou, Jack Jansen, Irene Viola, Shishir Subramanyam, and Pablo Cesar. 2021. CWIPC-SXR: Point Cloud Dynamic Human Dataset for Social XR. In *Proceedings of the 12th ACM Multimedia Systems Conference*.

[31] Nitish Srivastava, Elman Mansimov, and Ruslan Salakhudinov. 2015. Unsupervised Learning of Video Representations using Lstms. In *Proceedings of the International Conference on Machine Learning*.

[32] Yan Tian, Yujie Zhang, Wei-Gang Chen, Dongsheng Liu, Huiyan Wang, Huayi Xu, Jianfeng Han, and Yiwen Ge. 2022. 3D Tooth Instance Segmentation Learning Objectness and Affinity in Point Cloud. *ACM Transactions on Multimedia Computing, Communications, and Applications* (2022).

[33] Antonio W Vieira, Erickson R Nascimento, Gabriel L Oliveira, Zicheng Liu, and Mario FM Campos. 2012. Stop: Space-time Occupancy Patterns for 3d Action Recognition from Depth Map Sequences. In *Progress in Pattern Recognition, Image Analysis, Computer Vision, and Applications - 17th Iberoamerican Congress*.

[34] Irene Viola, Jelmer Mulder, Francesca De Simone, and Pablo Cesar. 2019. Temporal Interpolation of Dynamic Digital Humans Using Convolutional Neural Networks. In *Proceedings of the IEEE International Conference on Artificial Intelligence and Virtual Reality*.

[35] Guangming Wang, Xinrui Wu, Zhe Liu, and Hesheng Wang. 2021. Hierarchical Attention Learning of Scene Flow in 3d Point Clouds. *IEEE Transactions on Image Processing* (2021).

[36] Haiyan Wang, Jiahao Pang, Muhammad A Lodhi, Yingli Tian, and Dong Tian. 2021. FESTA: Flow Estimation via Spatial-Temporal Attention for Scene Point Clouds. In *Proceedings of the IEEE Conference on Computer Vision and Pattern Recognition*.

[37] Jun Wang, Xiaolong Li, Alan Sullivan, Lynn Abbott, and Siheng Chen. 2022. PointMotionNet: Point-Wise Motion Learning for Large-Scale LiDAR Point Clouds Sequences. In *Proceedings of the IEEE Conference on Computer Vision and Pattern Recognition*.

[38] Peng-Shuai Wang, Yang Liu, Yu-Xiao Guo, Chun-Yu Sun, and Xin Tong. 2017. O-CNN: Octree-Based Convolutional Neural Networks for 3D Shape Analysis. *ACM Transactions on Graphics* (2017).

[39] Xu Wang, Yi Jin, Yigang Cen, Tao Wang, and Yidong Li. 2021. Attention Models for Point Clouds in Deep Learning: a Survey. *arXiv preprint arXiv:2102.10788* (2021).

[40] Zirui Wang, Shuda Li, Henry Howard-Jenkins, Victor Prisacariu, and Min Chen. 2020. Flownet3d++: Geometric Losses for Deep Scene Flow Estimation. In *Proceedings of the IEEE Conference on Computer Vision and Pattern Recognition*.

[41] Mingqiang Wei, Zeyong Wei, Haoran Zhou, Fei Hu, Huajian Si, Zhilei Chen, Zhe Zhu, Jingbo Qiu, Xuefeng Yan, Yanwen Guo, et al. 2023. Agconv: Adaptive Graph Convolution on 3D Point Clouds. *IEEE Transactions on Pattern Analysis and Machine Intelligence* (2023).

[42] Yi Wei, Ziyi Wang, Yongming Rao, Jiwen Lu, and Jie Zhou. 2021. PV-RAFT: Point-voxel correlation Fields for Scene Flow Estimation of Point Clouds. In *Proceedings of the IEEE Conference on Computer Vision and Pattern Recognition*.

[43] Cheng Wencan and Jong Hwan Ko. 2021. Segmentation of points in the future: Joint segmentation and prediction of a point cloud. *IEEE Access* (2021).

[44] Xinshuo Weng, Jianren Wang, Sergey Levine, Kris Kitani, and Nicholas Rhinehart. 2020. Sequential Forecasting of 100,000 Points. *arXiv preprint arXiv:2003.08376* (2020).

[45] Wenxuan Wu, Zhi Yuan Wang, Zhuwen Li, Wei Liu, and Li Fuxin. 2020. Pointpwc-net: Cost volume on point clouds for (self-) supervised scene flow estimation. In *Proceedings of the European Conference on Computer Vision,*.

[46] Zonghan Wu, Shirui Pan, Fengwen Chen, Guodong Long, Chengqi Zhang, and S Yu Philip. 2020. A Comprehensive Survey on Graph Neural Networks. *IEEE Transactions on Neural Networks and Learning Systems* (2020).

[47] Chaoyun Zhang, Marco Fiore, Iain Murray, and Paul Patras. 2021. Cloudlstm: A Recurrent Neural Model for Spatiotemporal Point Cloud Stream Forecasting. In *Proceedings of the AAAI Conference on Artificial Intelligence*.

[48] Yi Zhang, Yuwen Ye, Zhiyu Xiang, and Jiaqi Gu. 2020. SDP-Net: Scene Flow Based Real-Time Object Detection and Prediction from Sequential 3D Point Clouds. In *Proceedings of the Asian Conference on Computer Vision*.

[49] Guanyu Zhu, Yong Zhou, Rui Yao, Hancheng Zhu, and Jiaqi Zhao. 2022. Cyclic Self-Attention for Point Cloud Recognition. *ACM Transactions on Multimedia Computing, Communications, and Applications* (2022).

[50] Wenjie Zhu, Zhan Ma, Yiling Xu, Li Li, and Zhu Li. 2020. View-dependent Dynamic Point Cloud Compression. *IEEE Transactions on Circuits and Systems for Video Technology* (2020).